\documentclass[11pt,a4paper]{article}
\usepackage{graphicx}
\usepackage{lscape}

\usepackage[hyperref]{emnlp2020}
\usepackage{times}
\usepackage{latexsym}

\usepackage{subcaption}

\usepackage{microtype}

\aclfinalcopy 
\def\aclpaperid{36} 

\usepackage{todonotes}
\newcommand{\Note}[2]{} 
\newcommand{\SideNote}[2]{} 
\renewcommand{\Note}[2]{\todo[color=#1,size=\small, inline=true]{#2}} 
\renewcommand{\SideNote}[2]{\todo[color=#1,size=\small]{#2}} %

\title{Scaling Systematic Literature Reviews with Machine Learning Pipelines}

\author{Seraphina Goldfarb-Tarrant\thanks{\hspace{0.2cm}Equal contribution, order determined by coin flip.} \and Alexander Robertson$^*$ \\
School of Informatics\\ University of Edinburgh \\
\And
Jasmina Lazic\\
Bayes Centre\\
University of Edinburgh \\
{\tt \{s.tarrant,alexander.robertson,jasmina.lazic\}@ed.ac.uk~~~~~~~~~~~~~~~~~~~~~~~~~~~~~~~}\\\AND
Theodora Tsouloufi \and Louise Donnison \and Karen Smyth \\
Supporting Evidence Based Interventions\\
Royal (Dick) School of Veterinary Studies\\
University of Edinburgh\\
{\tt \{theodora.tsouloufi,louise.donnison,karen.smyth\}@ed.ac.uk}

}

\date{today}

\begin{document}
\maketitle
\begin{abstract}
Systematic reviews, which entail the extraction of data from large numbers of scientific documents, are an ideal avenue for the application of machine learning. They are vital to many fields of science and philanthropy, but are very time-consuming and require experts. Yet the three main stages of a systematic review are easily done automatically: searching for documents can be done via APIs and scrapers, selection of relevant documents can be done via binary classification, and extraction of data can be done via sequence-labelling classification. Despite the promise of automation for this field, little research exists that examines the various ways to automate each of these tasks. We construct a pipeline that automates each of these aspects, and experiment with many human-time vs. system quality trade-offs. We test the ability of classifiers to work well on small amounts of data and to generalise to data from countries not represented in the training data. We test different types of data extraction with varying difficulty in annotation, and five different neural architectures to do the extraction. We find that we can get surprising accuracy and generalisability of the whole pipeline system with only 2 weeks of human-expert annotation, which is only 15\% of the time it takes to do the whole review manually and can be repeated and extended to new data with no additional effort.\footnote{\hspace{0.1cm}Code and links to models available at \url{https://github.com/seraphinatarrant/systematic_reviews}}
\end{abstract}

\section{Introduction}

Systematic reviews are part of the field of evidence-based analysis, and are a methodology for conducting literature surveys, where the focus is on comprehensively summarising and synthesising existing research for the purpose of answering research questions \cite{higgins2019cochrane}. The aim of this process is to be very broad coverage to avoid unknown bias creeping into results via the alternative of cherry-picking scientific results \cite{chalmers1995systematic}.
Conducting systematic reviews requires trained researchers with domain knowledge. The stages of the process are time-consuming, but vary in how much physical and mental labour they require \cite{borah2017analysis}. As a result, systematic reviews suffer from three primary challenges \cite{allen1999estimating, shojania2007quickly}:
\begin{enumerate}
\vspace{-.1em}
\setlength\itemsep{-.1em}
    \item they are very expensive, as they require many months of expert human labour;
    \item they easily become out of date, for the same reason;
    \item there is no amortised cost to human time at expanding them; human effort is linear in amount of research reviewed.
\end{enumerate}
So though systematic reviews have been shown to be very effective and less prone to human biases \cite{mulrow1994systematic}, these issues often prove prohibitive. \\ 
However, these challenges are well suited to Machine Learning solutions, and there has recently been an increase in interest in applying NLP to this process \cite{marshall2019toward}. In this paper, we investigate the feasibility of implementing the multi-stage human process of a systematic review as a Machine Learning pipeline. We construct a systematic review pipeline which aims to assist researchers and organisations focusing on livestock health in various African countries who previously performed reviews manually (via a process visualised in Figure \ref{fig:human_pipeline}). The pipeline begins with scraping for articles, then classifies them into whether or not to include in the review, then identifies data to extract and outputs a spreadsheet. We discuss the technical options we evaluated at each steps. Pipeline components are evaluated with intrinsic metrics as well as more pragmatic, extrinsic, considerations such as time and effort saved.

While previous work exists surveying the applicability of various Machine Learning methods and toolkits to the systematic review process (Section \ref{sec:rel_work}) and a few apply them, there are no extant studies that implement a full system and analyse the trade-offs between different methods of training data creation, different annotation schemas, human expert hours needed to build a system, and final accuracy. We experiment with all of these factors, as well as with a few different architectures, with the aim of informing the planning and implementation of systematic review automation more broadly.  

To further this goal, we particularly experiment with low resource scenarios and with generalisability. We investigate different thresholds for training data for the document classifier and different annotation schemas for the data extraction. We additionally test the ability of the system to generalise to documents from new countries.  

Key research questions are as follows:
\paragraph{Extraction} Which techniques are best for identifying and extracting the desired information?
\paragraph{Data Requirements} How much labelled training data is needed? Can existing resources be leveraged?
\paragraph{Re-usability} How generalisable is a pipeline to new diseases and countries?
\paragraph{Performance} What is the trade-off between pipeline accuracy and human time savings?
\paragraph{Architecture \& Pre-training} How important is model architecture as applied to extraction tasks? How important is embedding pre-training, and how important is pre-training on scientific literature vs. general content (domain match)?\\

We find that surprisingly little training data (and few human hours) are necessary to get an accurate document classifier, and that it generalises well to unseen African countries (Section~\ref{sec:classifier_results}), which enables systematic reviews to be expanded to new areas with essentially constant time. In our text extraction experiments, we find that both sentence and phrase level extraction models can each play a role in such a pipeline, 
but that phrase extraction, which has not previously been done for this task, performed better than expected both with baseline CNN models \cite{Yang2016HierarchicalAN} and with BERT-based Transformers \cite{Devlin2019BERTPO}, with Transformers based on scientific pre-training \cite{Beltagy2019SciBERTPC} performing best. We demonstrate how the creation of labelled training data can be sped up through annotation tools, and that consideration should be given to the balance of training examples present within this data, since doing so may require less data overall while still maintaining good performance. Furthermore, besides automatic information extraction, much labour in constructing systematic reviews can be saved through simply automating the process of searching and downloading documents. 

We empirically demonstrate that most of the three month pipeline of a systematic review can be automated to require very little human intervention, with acceptable accuracy of results. We release our code, annotation schema, and labelled data to assist in the expansion of systematic reviews via automation.

While we demonstrate this system on one domain, the framework is domain independent and could be applied to other kinds of systematic reviews. New training data and annotation schemes would be necessary to switch to medical or other domains, but our findings on time saving processes for annotation  would apply, and confidence thresholds that we implement are adjustable to customise to different levels of accuracy to human time trade-offs that are appropriate to different fields. Our exploration into necessary amounts of training data for accuracy and generalisability are broadly applicable.

\section{Background and Motivation}

As a case study, we work with the Supporting Evidence Based Interventions team at the Royal (Dick) School of Veterinary Studies at the University of Edinburgh, focusing on putting data and evidence at the centre of livestock decision-making in low and middle-income countries, predominantly in Africa. In these countries, livestock offer a path out of poverty for millions of smallholders, as well as providing vital nutrition for families and communities. While the veterinary technology and techniques required to improve livestock outcomes already exist (and are readily available to large scale commercial concerns worldwide), there is a lack of reliable information on animal health and productivity in these countries, at this scale. This data is needed not only in order to best target interventions, but to select the most efficient intervention in any particular context. 

There is very little data in this area for these countries, and it is often out of date. One proxy for direct measurement of livestock health in all herds in a country is the evidence found in veterinary science research publications, which have conducted prevalence studies. Individually, these studies give an indication of the prevalence of a specific disease in a specific region of a country at a specific time, affecting a specific breed of animal. But collectively, they give a much broader understanding of livestock health.

Four strands of data are of key importance: general livestock statistics (herd size and characteristics), health (mortality and disease), production (yields and growth rates) and economics (breeding costs, feeding costs, produce sale values). Here, we focus on health, specifically the prevalence of a wide range of diseases (e.g. brucellosis, foot and mouth disease) that effect ruminants (sheep, goats and cattle), with a focus on countries such as Nigeria, Ethiopia and Tanzania. 

\begin{figure}[ht]
\centering
\includegraphics[width=\linewidth]{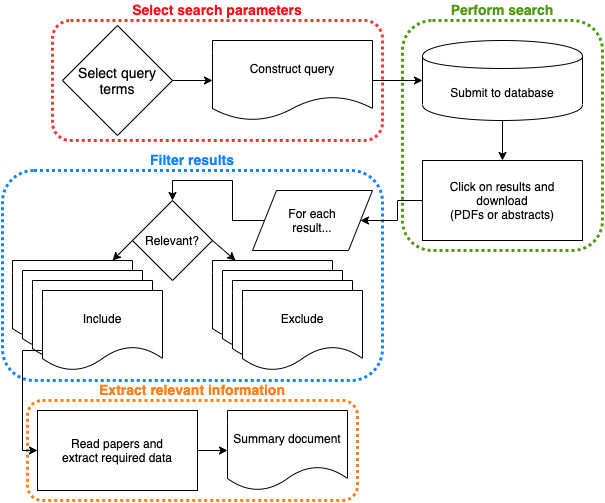}
\caption{Human-based information extraction systematic review pipeline.}
\label{fig:human_pipeline}
\vspace{-1em}
\end{figure}

However, collecting and summarising the findings of these studies is time-consuming manual work. For example, searching databases such as Google Scholar, Pubmed and Web of Science for prevalence studies, conducted between 2010 and 2018 on 28 diseases in small ruminants in Ethiopia, returned many thousands of results. Of these, 403 papers were considered relevant and information was extracted by experienced veterinarian researchers. The completely manual process, outlined in Figure~\ref{fig:human_pipeline}, produced high quality data but took approximately three months. 

The target information consists of dates (the start/end of the study), numerical data (sample size, direct/percentage prevalence numbers, animal age) and small lexical terms. These terms can be veterinary (diagnostic tests used, production systems, study design, statistical analysis performed, species) or geographic (region, ecosystem).\footnote{Appendix \ref{app:target_data} contains example target extracted data}

Since this summarisation process isn't abstractive, readers find it easy, if time-consuming, to identify the relevant information in research papers. The current project aimed to automate as much of the process as possible, to allow it to scale to a wider range of diseases (approximately 50) and countries. All but the initial component of Figure~\ref{fig:human_pipeline} can be entirely automated -- humans are still required to define search terms. 

\section{Pipeline Details \& Experimental Setup}
In the following we detail each stage of the constructed pipeline, how it compares to the human version and the human time saved. In the succeeding Section~\ref{sec:experiments} we detail metrics and evaluation for our targeted experiments with the classification and extraction pipeline components. 

\subsection{Document Search}
The highest return on investment in terms of engineering effort vs. human time-savings was our automation of literature searches. Previously, researchers would construct a list of searches to perform, input these into various databases, then manually download the results they considered relevant. An example search is {\tt (Livestock OR ruminants OR sheep OR goats OR cattle OR cow OR ram OR ewe OR bull) AND (Ethiopia) AND (Anthrax OR "Bacillus anthracis") AND (prevalence OR incidence)}

APIs are available for Scopus, Pubmed and Web of Science, for which we obtained institutional access. Google Scholar has no available API, so we used the SerpApi service\footnote{\url{https://serpapi.com/}} which provides paid programmatic access. The pipeline therefore maintains full coverage of paper sources.
As the APIs return only links to papers, PDFs still need to be retrieved. Issues to navigate here included links to websites rather than files, which requires additional negotiation through user-agent strings, parsing HTML for links to the PDF or parsing page headers to extract metadata redirecting to a PDF such as {\tt citation\_pdf\_url}. A generic approach was successful in most cases but site-specific downloaders had to be constructed for 38 domains, based on trial and error. 

Time spent on retrieval of potentially relevant documents to include in the systematic reviews were reduced dramatically, with the main limiting factor being rate throttles on APIs. Conducting searches on all four databases for 50 diseases in 3 countries takes approximately one hour on one machine and requires no human input. Parallel downloading of PDFs is even faster. This can be repeated at any interval to keep results up to date. By contrast, this step of the process used to take human experts 83 hours (2 weeks full-time) each time it was done, while covering fewer diseases and only one country.

\subsection{Document Classification}
Once search results have been collected, retrieved PDFs must be classified for inclusion vs. exclusion in the systematic review. Human reviewers use various criteria to assess inclusion (peer-review, type of study/experiment, subject matter) which, as we observed in user studies, they determine entirely from the title and abstract. 
We use the PDF DOI or ISSN to retrieve the title and abstract
\footnote{PDFs are converted to text in the following step and we could use the title and abstract from extraction, but PDF extraction is noisy and so we chose to rely on reference database lookups where available.} 
using the Wikimedia Citoid API
\footnote{\url{https://en.wikipedia.org/api/rest_v1}}. We then train an SVM Classifier implemented in Sci-kit Learn \cite{scikit-learn} with concatenated TF-IDF Vectors, which can train in under an hour on a standard Linux machine.
The classification process, which previously took a human expert 20 hours for 1000 documents, now generates results in minutes. We additionally implemented human review of documents with low classifier confidence (further details in Section \ref{sec:classifier_results}) upon consultation with our systematic review experts, as this both increases classifier accuracy and human trust in results. 

\subsection{PDF to Text Conversion}
Relevant PDFs are converted to plain text with the pdfminer.six package\footnote{\url{https://github.com/pdfminer/pdfminer.six}}. The text data extracted from PDFs can be noisy -- tables are especially problematic, headers/footers may end up inside main text, word and line spacing may be inconsistent, fonts may be improperly converted to text. The bulk of this can be overcome through basic pre-processing.

Once converted, the text is split into paper sections (e.g. abstract, introduction, methods) using regular expressions derived from manual inspection of 100 papers. This involves matching spans of text which appear between common section titles. For example, the abstract generally appears between `abstract' and `introduction', `abstract' and `keywords', `summary' and `introduction'.\footnote{PDF processing is documented fully in released code.}

\subsection{Data Extraction}
The goal of a systematic review is to output a tabular file where each column stores target information for each paper; this will then later be used to generate visualisations. Manually extracting this information is easy for knowledgeable humans: it isn't abstractive and does not require close reading of the full text. However, it is time-consuming at scale and does require experts, so both performing the process manually and creating training data incurs a significant cost. In addition, the different kinds of target information pose different technical challenges. Consider the sentence \textit{Rose Bengal Plate Test found 1.72\% (5/291) of the samples to be sero-positive}, which contains information about the diagnostic test used, the prevalence rate and sample size, all of which we want to capture. The phrase associated with the diagnostic test can be understood out of context but numbers generally cannot. Simply extracting all percentages from a text will be uninformative, rendering rule-based extraction approaches unsuitable. We therefore explored two machine learning approaches to automatic extraction to balance the difficulty in creation of annotated training data with suitability of the extraction approach.

A \textit{sentence-based} classifier can be used to label sentences as containing target information, and has been the method of extraction used in previous work \cite{marshall-etal-2017-automating, Kiritchenko2010ExaCTAE, Schmidt2020DataMI}. But this does not fit in well with the desired tabular output for all target fields: the same information can appear in multiple sentences and the same sentence can contain multiple targets. However, this approach is much easier for a human annotator. It should also work well for numerical targets: the context is preserved in the output and non-relevant numerical targets will be ignored or scored low. Alternatively, a \textit{phrase-based} classifier can apply labels to individual words and phrases within sentences. The extracted information will be more focused and should work well for phrase-based targets. The results will not require rule-based and human post-processing, as with the results of sentence-based extraction, but training data creation is more onerous. So given a fixed amount of human expert hours available, this approach may be less desirable, since it will generate much less training data.

We test the difference between both approaches using CNN-based text classification and named entity recognition models implemented in Prodigy\footnote{\url{https://prodi.gy/}}. This tool combines data annotation and model training. We created an annotation schema with 16 labels taken from manually created gold standard systematic review output.\footnote{Annotation schema included in code repository.} 

Creating training data for a sentence-based classifier is mechanically simple: the Prodigy annotation tool allows non-technical users to quickly assign labels using an interface with keyboard shortcuts, and we can display one sentence at a time.  Prodigy also allows phrase-level labelling, but this is a more involved process as the user must mark the start/end boundaries of a span and then apply the appropriate label. A single veterinarian labelled 4600 items at the sentence level in 56 hours, reporting the process to be easy and straightforward. The same veterinarian labelled 4200 items at the phrase level in 70 hours, reporting it to require much more physical and mental effort.

\section{Methodology}
\label{sec:experiments}
We performed detailed evaluation of the different classification and extraction components. 

\paragraph{Document Classification} We investigate the trade-off between training data volume and performance, and how generalisable a model is. For training volume, we fix a test set and reduce training data in chunks of 20\% of total. We test generalisability by training models on country-specific data and evaluating on unseen data from other countries. We report Accuracy overall, as well as Precision, Recall, and F1 on the \textit{include} label in this binary classification task. Finally, we investigate the effect of thresholding classifier confidence, and sending low confidence documents for manual human review, on both the accuracy of the system and on human time cost.

\paragraph{Data Extraction} We evaluate the sentence classifier and sequence-labelling approach with our CNN models. 
We also consider the impact of using document representations constructed with embeddings trained entirely on the source data, versus general purpose GloVe embeddings \cite{pennington2014glove} trained on web data, versus general purpose GloVe embeddings fine-tuned on the source data\footnote{Implemented in spacy \url{https://spacy.io/}}. As the sentence-level classifier is multi-label and multi-class, we report AUC (Area Under Curve).\footnote{AUC \= true positive vs false positive rate over a range of discrimination thresholds}
For the phrase-level sequence labelling approach, we report F1 score.

\subsection{Training Data Creation}

\begin{table*}[t]
\centering
\resizebox{0.8\textwidth}{!}{%
\begin{tabular}{rrcc}
\hline
\multicolumn{1}{c}{Data} & \multicolumn{1}{c}{Description} & Phrases & Sentences \\ \hline
disease & Animal disease & 3307 (31.5\%) & 778 (31.4\%) \\
species & Species studied & 2002 (19.1\%) & 518 (20.9\%) \\
region & Area within country & 1487 (14.2\%) & 298 (12.0\%) \\
individual\_prevalence & Number of infected animals & 743 (7.1\%) & 172 (6.9\%) \\
diagnostic\_test & Test used to detect disease & 729 (6.9\%) & 172 (6.9\%) \\
reference & Reference to another study & 591 (5.6\%) & 137 (5.5\%) \\
sample\_type & Biological samples used & 486 (4.6\%) & 117 (4.7\%) \\
statistical\_analysis & Analysis performed & 261 (2.5\%) & 63 (2.5\%) \\
age & Ages of animals tested & 228 (2.2\%) & 65 (2.6\%) \\
sample\_size & Number of animals tested & 161 (1.5\%) & 24 (1.0\%) \\
production\_system & Type of farm & 141 (1.3\%) & 43 (1.7\%) \\
ecosystem & Geography of farm & 141 (1.3\%) & 44 (1.8\%) \\
study\_design & Type of study used & 120 (1.1\%) & 28 (1.1\%) \\
study\_date & Date study was conducted & 64 (0.6\%) & 8 (0.3\%) \\
herd\_prevalence & Number of herds infected & 28 (0.3\%) & 7 (0.3\%) \\
mortality & Animals killed by disease & 5 (0.0\%) & 1 (0.0\%) \\ \hline
\end{tabular}%
}
\caption{Proportion of target items identified during data annotation.}
\label{table:tags}
\vspace{-1em}
\end{table*}

\paragraph{Document Classification} veterinarian experts manually labelled papers as include/exclude: 608 papers from searches for 50 diseases for the countries of Ethiopia, Nigeria and Tanzania. We experimented with labelling 100 test documents: half via a reference manager/document reader\footnote{Zotero (\url{zotero.org}) was chosen as it is the only service which provides an API} and via a simple spreadsheet interface where one column contained the paper title, one contained the paper abstract, and the expert filled in a third column for the include/exclude label.\footnote{Recall that criteria for inclusion in the study are fully determinable via these fields.} The spreadsheet method was 3 times faster than using a reference manager, enabling experts to complete the 608 papers of training data in 5 hours. Half the data contains country information, so we use only that half for our generalisability experiments.

\paragraph{Data Extraction} 52 documents were randomly sampled from the set of documents manually classified for inclusion. The sampled documents covered 13 diseases for studies in Ethiopia, Nigeria and Tanzania. Only abstracts, results and methods sections were annotated.

To select a manageable volume of data for annotation, and avoid including noisy data from the PDF extraction process, we applied some restrictions. For the sentence-based task, all sentences of at least 9 words within the abstract were included, along with a random sample of 150 sentences (between 9 and 25 words long) from the results and methods sections. Sentence length was based on the fact that very short/long sentences were generally noisy due to the PDF conversion process. For the phrase-based task, sections were split into chunks of three sentences to preserve some context. The entire abstract was used, plus a random sample of 25 chunks from each of the methods and results sections.

Table~\ref{table:tags} briefly describes each item and the breakdown of label frequency in our annotated data. There is a clear imbalance in label frequency -- some are not commonly reported in general (e.g. mortality, herd prevalence) while others are reported very few times per paper (e.g. study date).

\subsection{Experimental Conditions}
\paragraph{Data Volume} We trained document classification models using proportions from 20\% to 100\% of all data. 
\paragraph{Generalisability} Three document classification models were each trained on two of the three countries, with the final country held out. We included data volume ablations in these experiments as well.
\paragraph{Sentence vs. Phrase Models}
We trained the CNN-model on sentence labelled vs. phrase labelled data to assess the feasibility of using each annotation approach.
\paragraph{Architecture \& Pretraining}
We experiment with five different architectures for the phrase-based models. We use the Prodigy CNN with randomly-initialised embeddings, the Prodigy CNN with frozen pre-trained embeddings, the Prodigy CNN with pre-trained embeddings fine-tuned on our data, distilBERT \cite{Sanh2019DistilBERTAD}, and SciBERT \cite{Beltagy2019SciBERTPC}. The CNN is easy to implement out of the box, as it is built into the annotation tool, can be trained without access to a GPU, and could potentially be less data-hungry than a transformer - all important considerations in our resource constrained setting. Adding pre-trained embeddings allows us to isolate the effect of pre-training from the effect of architecture. Since the phrase-labelling task is well suited to the masked language modelling objective, we additionally experiment with fine-tuning distilBERT (which is reasonably sized for our small amount of data) and SciBERT, to test whether the domain match of pre-trained data matters.

\section{Results}
\label{sec:classifier_results}

\begin{figure*}[t]
\centering
\includegraphics[width=0.9\textwidth]{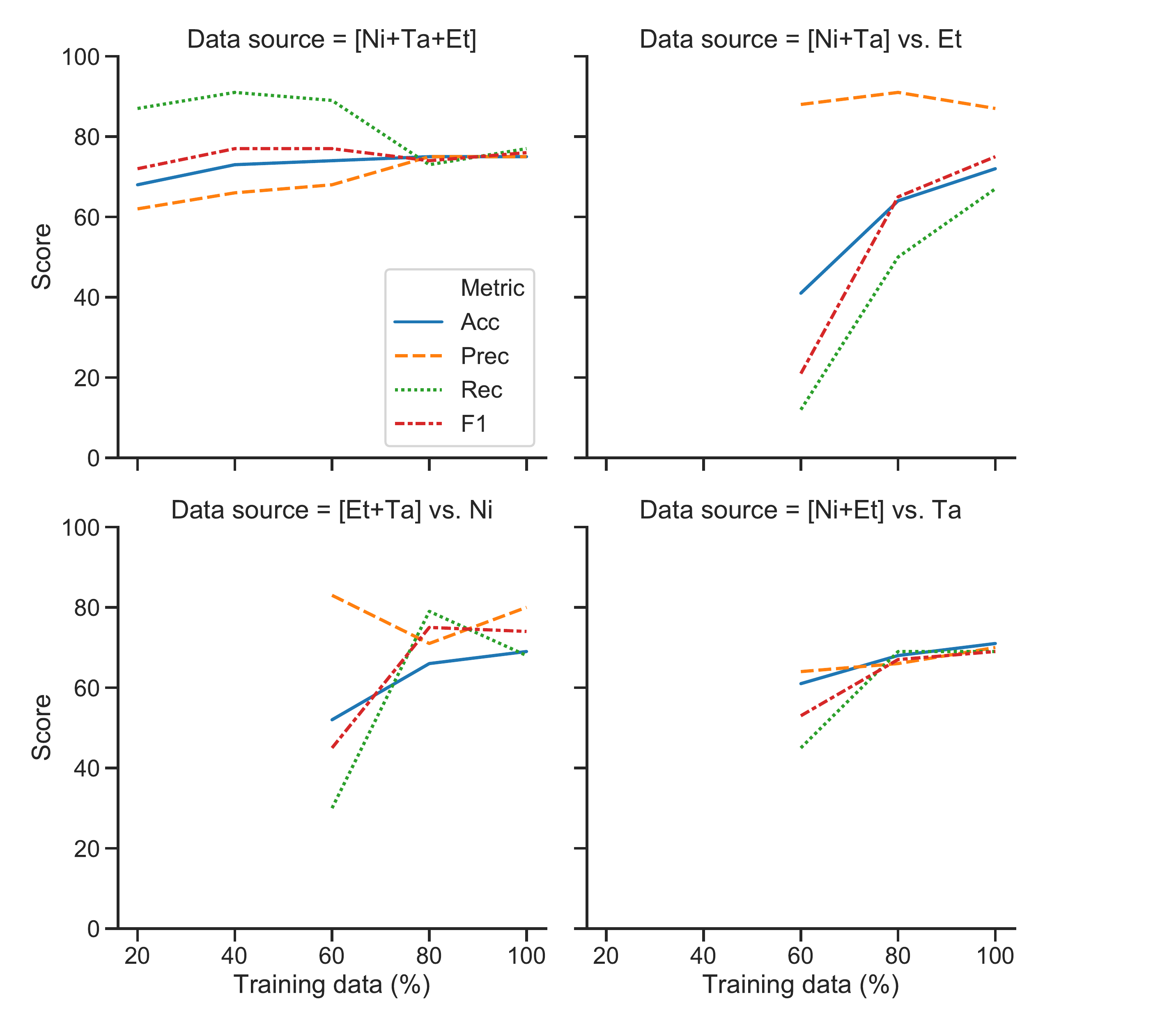}
\caption{Accuracy, Precision, Recall, and F1 for the document classification model showing performance changes as training data is decreased for the full set of 608 documents (85-15 train-test split), as well as generalisability to held out countries. Et=Ethiopia, Ta=Tanzania, and Ni=Nigeria. Note that country experiments have only 200 train and 100 test documents (100 per country, with test held fixed).}
\label{fig:doc_results}
\vspace{-1em}
\end{figure*}

Results for document classification experiments are shown in Figure~\ref{fig:doc_results}. 
The upper left quadrant of Figure~\ref{fig:doc_results} contains data for 608 documents with an 85-15 train-test split across all 3 countries, showing an expected increase in classifier performance as data increases, but levelling off slightly by 80\% of the full training volume. The other quadrants show the same data for 100 documents per country (200 train, 100 test) but with a minimum of 60\% of total data, as with less than 100 training samples the model does not converge.

For the data volume studies on the full dataset, a notable trend is that recall is quite high even with very little training data (~100 documents), and that what the classifier learns with additional data is predominantly a better precision-recall balance. For the held-out-country generalisability studies, the amount of training data is more important, and recall is no longer high immediately. 

This suggests that for a fixed country with a semi-automated system that has resources for a secondary human-filtering, very little training data is necessary. However, extensibility to new countries does require more data. Given that additional more data, performance on unseen countries is equal to that of known countries of equivalent training set volumes. This suggests an important new extensibility opportunity for systematic review systems.

In practice, our experts needed slightly higher accuracy than the best combined accuracy. To address this we implemented confidence-thresholding, such that documents below a user-set threshold are uploaded to a \textit{needs review} folder, which generates a weekly email. 15\% of test documents require review at our final confidence threshold, which reduces human time to ~20 min per 100 documents classified, but allows for an increased accuracy to 88\%. Human reviews are then fed back into classifier training, which should incrementally improve confidence and reduce human labour over time. We leave that longitudinal study for future work.  

\begin{figure*}[t]
  \centering
  \begin{subfigure}[t]{0.35\textwidth}
    \includegraphics[width=\textwidth]{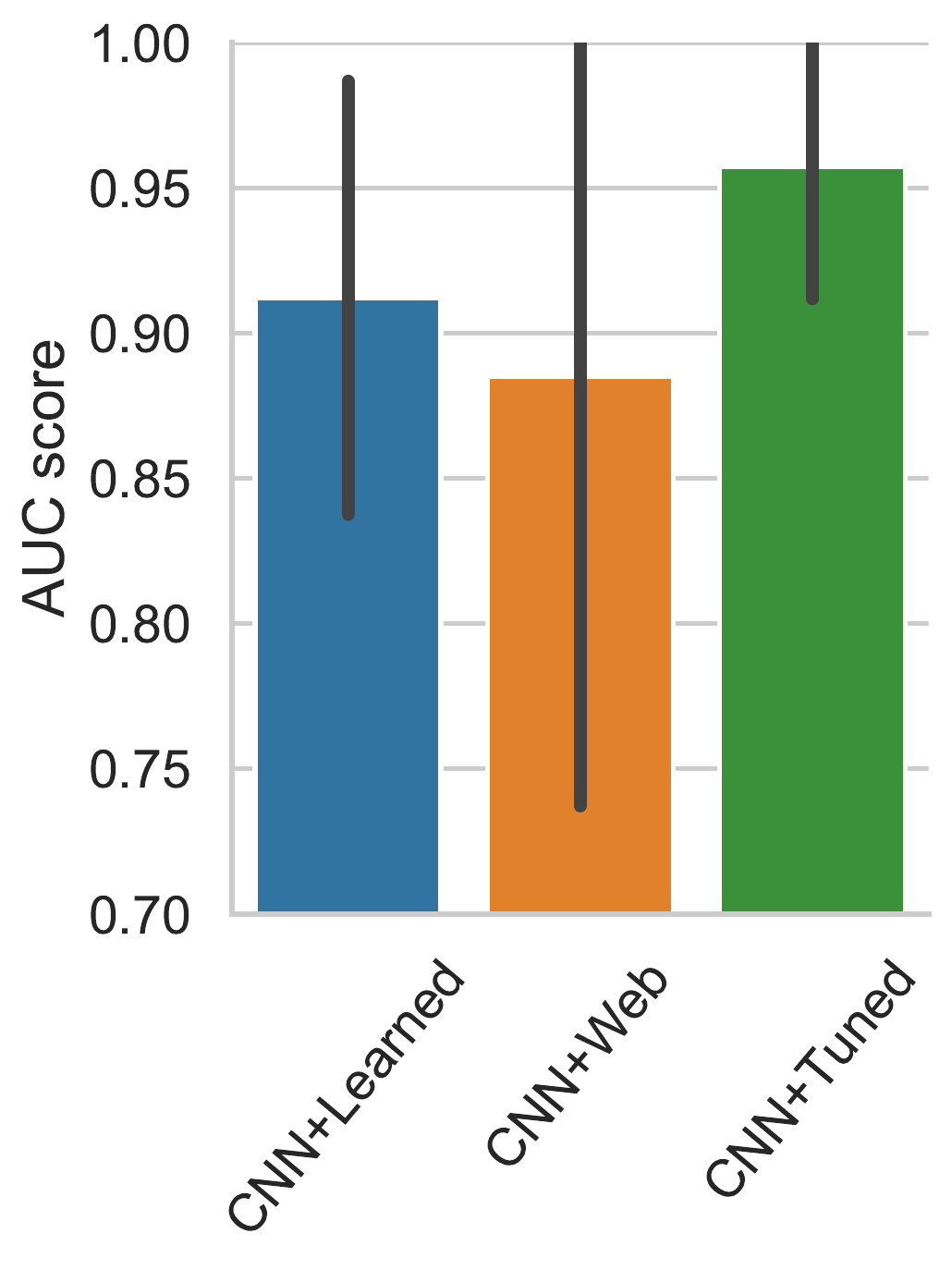}
    \caption{Sentence-level classification: Mean AUC score, over all labels, for three embedding sources. Error bars denote standard deviation.}
    \label{fig:mean_sent_results}
    \end{subfigure} \hfill
    \begin{subfigure}[t]{0.55\textwidth}
    \includegraphics[width=\textwidth]{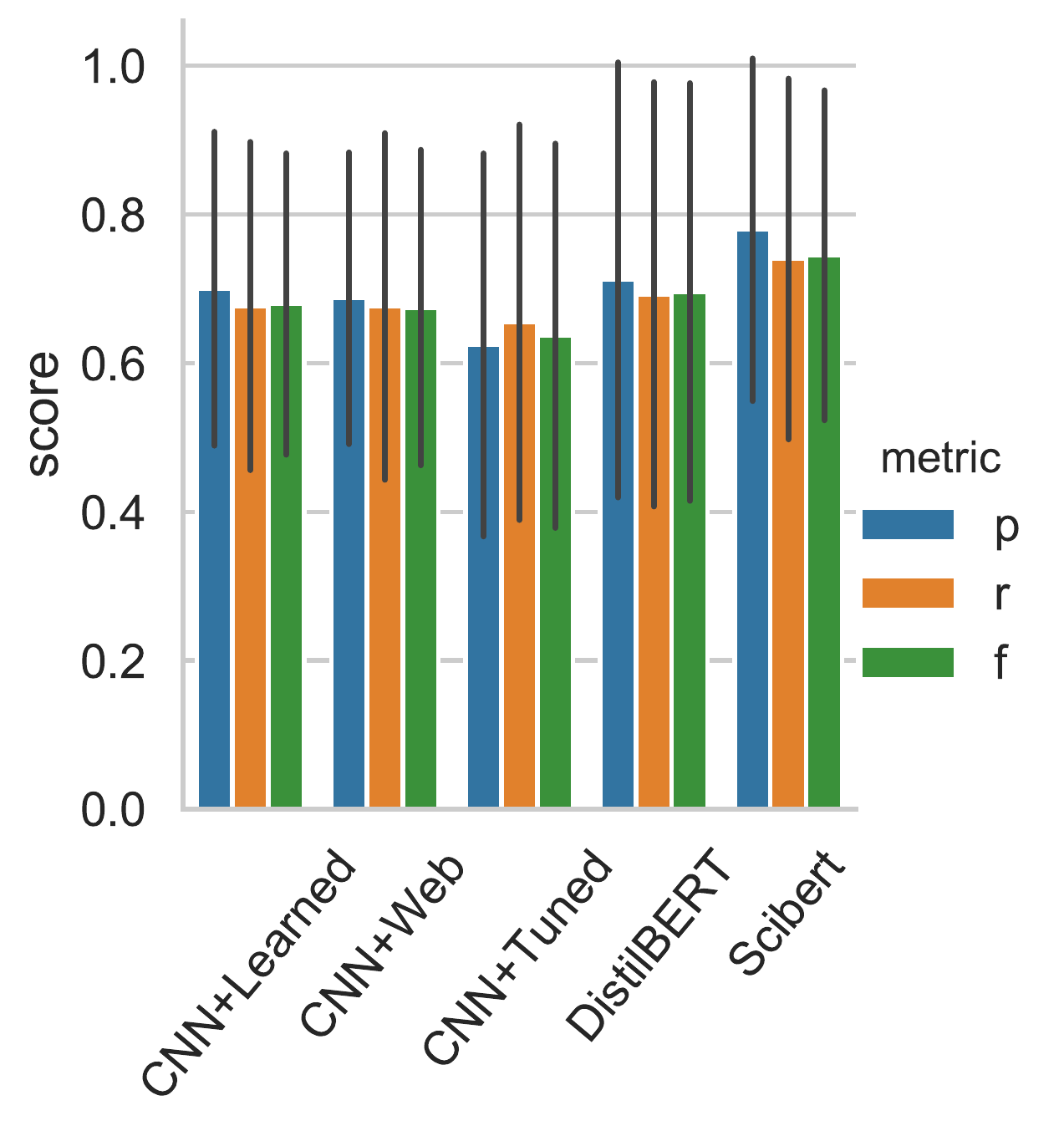}
    \caption{Phrase-level classification: Mean Precision/Recall/F1, over all labels, for three embedding sources and two Transformer-based models. Error bars denote standard deviation.}
    \label{fig:mean_phrase_results}
    \end{subfigure}
    \vspace{-1em}
\end{figure*}

Results for three CNN-based sentence-level data extraction models are shown in Figure~\ref{fig:mean_sent_results}. We report mean AUC score on all labels (with standard deviation shown) with an 80/20 train/test split.

For sentence-level models, fine-tuned web embeddings give better performance overall. Mean AUC score was 0.96 (stdev 0.05) using fine-tuned general purpose web embeddings; 0.91 (stdev 0.08) using learned embeddings; 0.89 (stdev 0.15) using general purpose web embeddings. 

For phrase-level models, this is no longer true: mean F1 score was 0.67 (stdev 0.22) using pre-trained general purpose web  embeddings; 0.68 (stdev 0.2) using learned embeddings; 0.64 (stdev 0.27) using fine-tuned general purpose web embeddings. Transformer-based models performed more strongly: F1 for DistillBERT was 0.70 (stdev 0.29), SciBERT 0.75 (stdev 0.23).

Focusing on the items considered most important by the veterinarian researchers (disease, species, region, individual prevalence, diagnostic test, sample type, sample size, study date), results in an increase of ~0.03 AUC for each sentence-level model. Phrase-based models F1 score increases by ~0.10. 

These results suggest that pre-training is important for the sentence-based classifier, and that the BERT-based Transformer architecture with the masked language modelling objective can do very well on phrase-level extraction and bring performance high enough to make this approach feasible. However, they show that domain-specific pretraining data has a larger effect than architectural differences. While Transformer-based models for phrase-level labelling out-performed CNN-based models, it was the SciBERT model trained on academic papers, then fine-tuned on our specific task, which gave the best performance, and a larger performance boost than the initial jump from BERT.
The best type of pre-training does vary based on type of extraction: general purpose embeddings perform worst for sentence-level labelling, though are on par with those learned from the training data for phrase-level labelling.

We analyse per-label performance for the SciBERT model to verify phrase-level feasibility, and include this data in Figure~\ref{fig:scibert_phrase_results} in Appendix \ref{sec:per_label}. Performance is generally high, even for some low-frequency items. Some of these were uncommon in our training data (due to appearing only once or twice per paper) but naturally appear in \emph{many} academic papers in general, which goes towards accounting for the success of SciBERT on this task. For example, SciBERT was the only model to correctly identify \emph{any} instances of herd prevalence.



\section{Related Work}
\label{sec:rel_work}
The application of NLP to systematic reviews is relatively new, but has been recently receiving more attention. There is a growing body of work that assesses the potential for automation in systematic reviews, but little that builds systems for the purpose and tests them empirically.

\newcite{marshall2019toward} review available tools that can be used to automate each element of the systematic review pipeline. \newcite{Marshall2020SemiAutomatedES} further review opportunities for semi-automation and assess opportunities and risks. \newcite{marshall2015systematic, o2015using} conduct systematic reviews of automation for systematic reviews. \newcite{thomas2017living} analyse the systematic review pipeline to find ways that human-machine collaboration can be applied and improve the speed.  

\newcite{marshall-etal-2017-automating} create a PDF viewer that humans can use to make the systematic review process easier and faster, by training a CNN to assess risk of bias in a document (an important part of evidence-based analysis in the medical domain, though not for our particular task) and identifies and displays sentences to the user that contain a subset of the information necessary for a systematic review. \newcite{Kiritchenko2010ExaCTAE} create an extraction system that identifies sentences and then post-processes them to extract data, but operate only on structured HTML \& XML. \newcite{Schmidt2020DataMI} apply fine-tuned BERT-based Transformers to the task of to sentence classification for semi-automated systematic review. \newcite{Goswami2019AnEM} build a PDF retrieval system for systematic reviews for psychology and use a random forest classifier to identify sentences for extraction.

As far as we are aware, no other work builds a phrase-based system, tests data volume and generalisability, or applies a diverse set of modern architectures to the task.

\section{Conclusion \& Future Work}
We investigated the application of automation to all stages of the systematic review pipeline for our veterinary research case study. We found that with two weeks (~80 hours) of human expert annotation we can automate a systematic review that previously took 3 months, and still maintain high levels of accuracy. Our classification system generalises well, enabling it to be applied to new countries for additional systematic reviews with no additional human annotation cost. Sentence-based and phase-based data extraction both perform well, and the creation of phrase-based training data can still fit within a small amount of human annotation hours and avoids the need for extensive post-processing. Fine-tuned BERT-based Transformers perform best at data extraction, with BERT pre-trained on scientific data giving the largest boost in performance, though a baseline CNN still performs surprisingly well. In future work, we plan to test generalisability cross-lingually, expand the generalisability tests to extraction as well as classification, and study the performance improvements of continuous training of classifiers on human corrections of low-confidence output.

\bibliographystyle{acl_natbib}
\bibliography{anthology,emnlp2020}

\clearpage

\appendix

\section{Target Extracted Data}
\label{app:target_data}
In Table~\ref{table:target_data} is the first 15 lines of a sample gold standard target data from a human systematic review (broken into two tables for display) that we use as a template for building our system. Note that some fields are blank because information was not found in or relevant to a given entry.

\begin{table*}[]
\centering
\resizebox{\textwidth}{!}{%
\begin{tabular}{llllllllllll}
ROW\_NUMBER & IDENTIFIER & YEAR\_PUBLICATION & REFERENCE & START\_DATE\_DATA & END\_DATE\_DATA & STATE & ECOSYSTEM & PRODUCTION\_SYSTEM & SPECIES & AGE & AGE\_DETAIL \\
1 & Nigussie et al; 2010 & 2010 & Nigussie et al & 2007 & 2008 & Oromia &  & Mixed farming & Cattle &  &  \\
2 & Regassa et al; 2010 & 2010 & Regassa et al & 2007 & 2008 & SNNPR &  &  & Cattle &  &  \\
2 & Regassa et al; 2010 & 2010 & Regassa et al & 2007 & 2008 & SNNPR &  &  & Cattle &  &  \\
3 & Regassa et al; 2010 & 2010 & Regassa et al & 2007 & 2008 & SNNPR &  &  & Cattle &  &  \\
4 & Bekele et al; 2010 & 2010 & Bekele et al & 2008 & 2003 & SNNPR &  &  & Cattle &  &  \\
5 & Shiferaw et al 2013; 2010 & 2010 & Shiferaw et al 2013 & 2007 & 2008 & Afar &  &  & Cattle &  &  \\
6 & Shiferaw et al 2011; 2010 & 2010 & Shiferaw et al 2011 & 2007 & 2008 & Afar &  &  & Cattle &  &  \\
7 & Shiferaw et al 2010; 2010 & 2010 & Shiferaw et al 2010 & 2007 & 2008 & Afar &  &  & Cattle &  &  \\
8 & Shiferaw et al 2014; 2010 & 2010 & Shiferaw et al 2014 & 2007 & 2008 & Afar &  &  & Cattle &  &  \\
9 & Shiferaw et al 2012; 2010 & 2010 & Shiferaw et al 2012 & 2007 & 2008 & Afar &  &  & Cattle &  &  \\
10 & Kumsa et al; 2010 & 2010 & Kumsa et al & 2006 & 2006 & SNNPR &  &  & Sheep &  &  \\
11 & Kumsa et al; 2010 & 2010 & Kumsa et al & 2006 & 2006 & SNNPR &  &  & Sheep &  &  \\
12 & Kumsa et al; 2010 & 2010 & Kumsa et al & 2006 & 2006 & SNNPR &  &  & Sheep &  &  \\
14 & Amenu et al; 2010 & 2010 & Amenu et al & 2007 & 2007 & Oromia &  & Mixed farming & Cattle &  &  \\
14 & Amenu et al; 2010 & 2010 & Amenu et al & 2007 & 2007 & Oromia &  & Mixed farming & Cattle &  & 
\end{tabular}%
}

\resizebox{\textwidth}{!}{%
\begin{tabular}{llllllllll}
\\
\hline
\\
DISEASE & SAMPLE & DIAGNOSTIC\_TEST & MEASUREMENT & NUMBER\_POSITIVE & NUMBER\_TESTED & PERCENTAGE & CALCULATION & COMMENTS & SOURCE \\
BVD & Serum & i-ELISA & Individual Prevalance & 65 & 567 & 11.4638447971781 & TOTAL & national surveillance/mixed altitudes (midland, highland)/zone\&sex splitting/adult\textgreater{}young(\textless{}3y) & LITERATURE \\
Tb & Intraderm  test & CIDT & Herd Prevalance & 19 & 39 & 48.7179487179487 & TOTAL & \textgreater{}6 MONTHS & LITERATURE \\
Tb & Intraderm  test & CIDT & Individual Prevalance & 48 & 413 & 11.6222760290557 & TOTAL & \textgreater{}6 MONTHS & LITERATURE \\
Tb & PM specimen & PM & Individual Prevalance & 11 & 1023 & 1.0752688172043 & TOTAL &  & LITERATURE \\
TRYPs & Blood & BC & Individual Prevalance & 71 & 323 & 22 & TOTAL & East African zebus, \textgreater{}1y/T. congolense, vivax\&brucei splitting & LITERATURE \\
FMD &  & Survey & Individual Mortality &  &  & 0.73 & TOTAL &  & LITERATURE \\
Pasteurelloses &  & Survey & Individual Mortality &  &  & 1.5 & TOTAL &  & LITERATURE \\
CBPP &  & Survey & Individual Mortality &  &  & 2.5 & TOTAL &  & LITERATURE \\
Blackleg &  & Survey & Individual Mortality &  &  & 0.13 &  &  & LITERATURE \\
Anthrax &  & Survey & Individual Mortality &  &  & 1.3 &  &  & LITERATURE \\
Endoparasites & Faeces & Floatation, Microscopy & Individual Prevalance &  &  & 6.7 & TOTAL & Strongyloides papillosus & LITERATURE \\
Endoparasites & Faeces & Floatation, Microscopy & Individual Prevalance &  &  & 15 & TOTAL & Trichuris spp & LITERATURE \\
Endoparasites & Faeces & Floatation, Microscopy & Individual Prevalance &  &  & 100 & TOTAL & Gi parasites/Strongyle eggs & LITERATURE \\
Tb & Intraderm  test & SCIDT & Herd Prevalance &  &  & 35 & TOTAL &  & LITERATURE \\
Tb & Intraderm  test & SCIDT & Individual Prevalance & 27 & 425 & 6.35 & TOTAL &  & LITERATURE
\end{tabular}%
}
\caption{Example target extracted data from a gold-standard human systematic review}
\label{table:target_data}
\end{table*}

\section{Detailed Analysis of Phrase-level Classification Performance}
\label{sec:per_label}
\begin{figure*}[t]
\centering
\includegraphics[width=\textwidth]{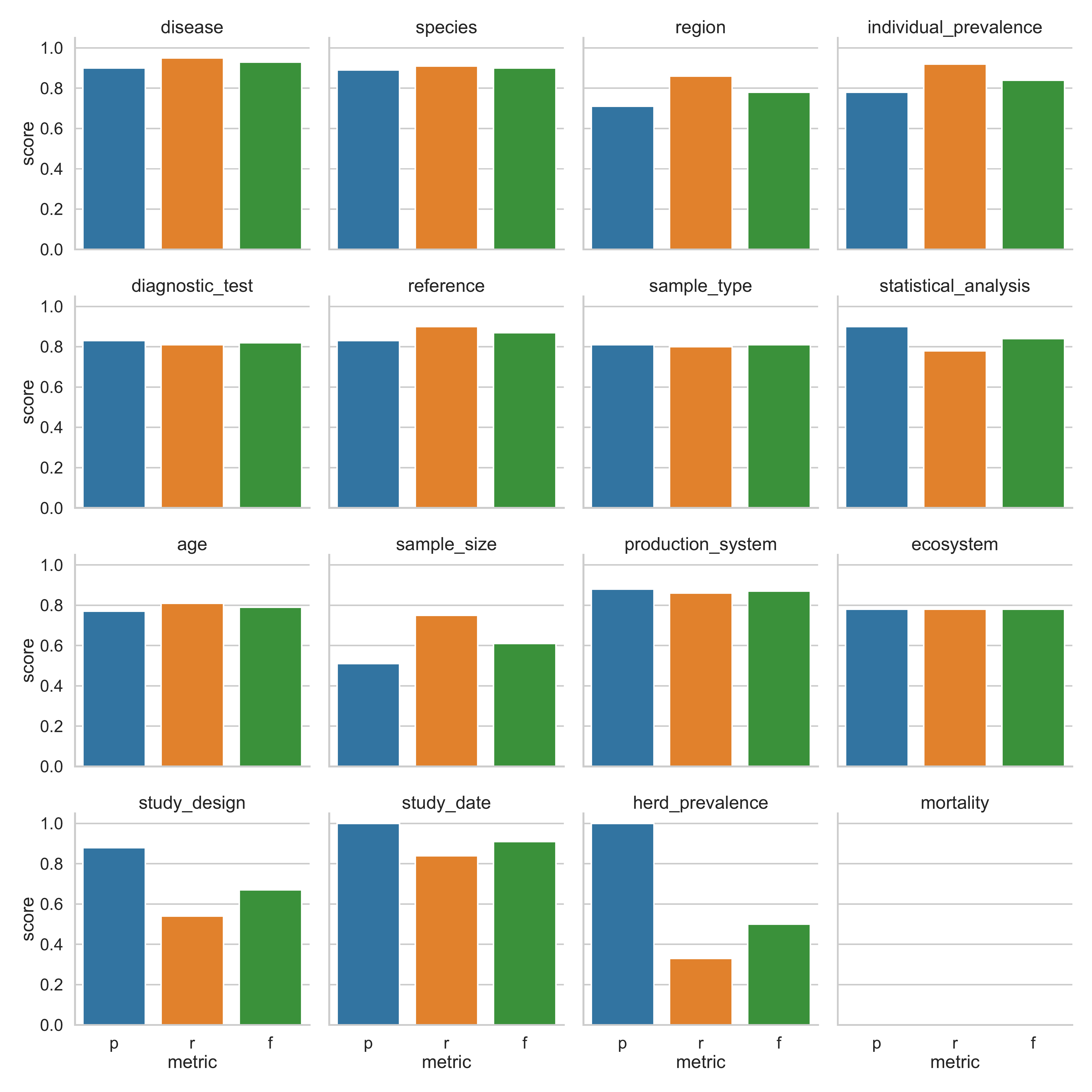}
\caption{Phrase-level classification: Precision/Recall/F1, per label, for SciBERT. From top left to bottom right: most to fewest examples in training data.}
\label{fig:scibert_phrase_results}
\end{figure*}

Displayed in Figure~\ref{fig:scibert_phrase_results} are the per label performance breakdowns for SciBERT, the strongest phrase-level extraction model. Performance remains high across many individual labels, with changes in performance mostly tracking with commonness of the information (and thus, how much training data is available for a fixed set of annotated documents). The exceptions to this trend are the \textit{region} and \textit{sample\_size} labels, which  have lower performance compared to equivalently common labels

\end{document}